\documentclass[conference]{IEEEtran}
\IEEEoverridecommandlockouts
\newif\ifblindreview
\blindreviewfalse      % <-- FALSE for camera-ready (authors visible)
\makeatletter\def\@IEEEcopyrightstring{}\makeatother
\usepackage[T1]{fontenc}
\usepackage[utf8]{inputenc}
\usepackage{amsmath,amssymb}
\usepackage{graphicx}
\usepackage{booktabs}
\usepackage{multirow}
\usepackage{array}
\usepackage{tabularx}            % tables that obey \columnwidth
\usepackage{siunitx}
\DeclareSIUnit\kph{km/h}
\DeclareSIUnit\mps{m/s}
\usepackage[section]{placeins}   % flush floats at section boundaries
\usepackage{dblfloatfix}         % improve figure* placement in 2-col
\usepackage{xcolor}
\usepackage{microtype}
\usepackage{balance}
\usepackage{cite}
\usepackage{url}
\usepackage[hidelinks]{hyperref}
\newcommand{\HDVtoAV}{\ensuremath{\mathrm{HDV}\!\to\!\mathrm{AV}}}
\newcommand{\HDVtoHDV}{\ensuremath{\mathrm{HDV}\!\to\!\mathrm{HDV}}}
\newcolumntype{L}{>{\raggedright\arraybackslash}X}
\newcolumntype{C}[1]{>{\centering\arraybackslash}p{#1}}
\begin{document}
% -- Title -------------------------------------------------------------
\title{Cut-In Gap Acceptance Toward Autonomous vs.\ Human-Driven
Vehicles: Evidence from the Waymo Open Motion Dataset}
% -- Authors -----------------------------------------------------------
\ifblindreview
  \author{
    \IEEEauthorblockN{Anonymous Authors}
    \IEEEauthorblockA{Anonymous affiliation\\Anonymous email}
  }
\else
  \author{%
    \IEEEauthorblockN{Abdulaziz Alhuraish, Yuhang Wang, Hao Zhou}
    \IEEEauthorblockA{Department of Civil and Environmental Engineering\\
    University of South Florida, Tampa, FL, USA\\
    \texttt{aalhuraish@usf.edu, yuhang.wang@usf.edu, haozhou1@usf.edu}}
  }
\fi
\maketitle
% ======================================================================
\begin{abstract}
Autonomous vehicles (AVs) are widely known to follow conservative, rule-based motion policies that surrounding drivers can learn to anticipate.  A direct consequence is that human drivers may accept mshorter longitudinal gaps when cutting in front of an AV than when targeting another human-driven vehicle (HDV).  We test this hypothesis using the Waymo Open Motion Dataset (WOMD), which provides 25{,}906 real-world highway scenarios at \SI{10}{\hertz}.  An eight-criterion lane-change detector extracts \textbf{706 \HDVtoAV} and \textbf{3{,}172 \HDVtoHDV} cut-in events from the same traffic environment.  The median accepted gap in front of the Waymo AV is \textbf{7.58\,m} versus \textbf{9.57\,m} for HDV targets---a \textbf{1.99\,m} reduction that is statistically significant ($p{=}5.76{\times}10^{-8}$, $d{=}{-}0.224$) and persists under speed-matched resampling.  Cut-in speeds toward the AV are 37\% higher (51.7 vs.\ 37.7\,km/h, $d{=}0.502$), and 68.0\% of AV-targeted cut-ins occur below the \SI{10}{\metre} gap boundary versus 51.8\% of HDV-targeted events ($\chi^2{=}60.5$, $p{<}10^{-13}$).  These results reveal a systematic and safety-relevant asymmetry in human gap-acceptance behavior that warrants AV-specific calibration of both motion-planning safety envelopes and traffic simulation models.
\end{abstract}


% ======================================================================
\section{Introduction}\label{sec:intro}

The deployment of autonomous vehicles (AVs) on public roads creates a
new form of mixed traffic in which human-driven vehicles (HDVs) and
AVs share infrastructure while operating under very different decision
processes.  AVs adhere to conservative collision-avoidance
policies---maintaining large headways, decelerating smoothly, and
never contesting gaps through the throttle or horn.  These properties
are widely understood by the public, raising a concrete safety concern:
do human drivers exploit this predictability by accepting shorter
longitudinal gaps when cutting in front of an AV than they would in
front of an equally attentive human driver?

Lane-change gap acceptance has been studied extensively in
HDV-to-HDV settings.  Naturalistic driving studies and large-scale
trajectory datasets have documented characteristic accepted-gap
distributions~\cite{yang2019,lee2004,wangXue2019}, and logit/probit
models have been calibrated
accordingly~\cite{daganzo1981,toledo2007}.
A growing body of literature reports that human behavior shifts in the
presence of an AV~\cite{soni2022,wen2022,rahmati2019}, yet no study
has directly compared \emph{cut-in gap acceptance} between AV-targeted
and HDV-targeted maneuvers using a large, naturalistic,
within-environment dataset.

We address this using the Waymo Open Motion Dataset
(WOMD)~\cite{womd2021}, which records real-world driving from the AV's
perspective while simultaneously tracking all surrounding agents.  This
architecture allows a direct, within-scenario comparison of cut-in
behavior toward the AV and toward nearby HDVs under identical traffic
conditions---a form of natural control that observational studies
rarely provide.

\smallskip\noindent\textbf{Contributions.}
\begin{enumerate}
  \item We design an eight-criterion, AV-centric cut-in detector that
        extracts both \HDVtoAV\ and \HDVtoHDV\ events from
        the same scenario recordings.
  \item We provide the first large-scale naturalistic comparison of
        cut-in gap acceptance ($n{=}706$ vs.\ $n{=}3{,}172$) across
        seven safety metrics spanning longitudinal, lateral, and
        kinematic dimensions.
  \item We quantify effect sizes and the practical safety implications
        of the observed asymmetry, offering concrete guidance for AV
        motion planning and traffic simulation calibration.
\end{enumerate}

% ======================================================================
\section{Related Work}\label{sec:related}

\subsection{Gap Acceptance in Lane Changes}

Gap acceptance---the binary decision to enter a longitudinal space in
an adjacent lane---has been modeled formally since
Daganzo~\cite{daganzo1981}.  Classical logit and probit formulations
link acceptance probability to gap size, relative speed, and driver
urgency~\cite{toledo2007}.  Yang et al.~\cite{yang2019} analyzed
3{,}000+ naturalistic lane changes and reported a median accepted gap
near \SI{8}{\metre} with a strong coupling to relative speed.  Using
the NHTSA 100-car Naturalistic Driving Study, Lee et
al.~\cite{lee2004} showed that rear-end near-crashes cluster when TTC
falls below \SI{2}{\second} during gap acceptance---a threshold now
embedded in many safety-assessment frameworks.  Wang et
al.~\cite{wangXue2019} further classified cut-in urgency via TTC and
deceleration rate required to avoid a crash (DRAC), identifying
approach speed as the primary risk discriminator.  These HDV-to-HDV
findings establish the baseline against which our AV-targeted
measurements are compared.

\subsection{Human Behavior Toward Autonomous Vehicles}

Human drivers are not passive observers of nearby AVs.  In a field
experiment, Soni et al.~\cite{soni2022} found smaller critical gaps
and shorter following headways when participants interacted with an AV,
attributing this explicitly to exploitation of the AV's defensive
policy.  Using Waymo Open Dataset car-following records, Wen et
al.~\cite{wen2022} documented shorter time headways and higher driving
volatility when the lead vehicle is an AV.  Rahmati et
al.~\cite{rahmati2019} measured statistically significant headway
reductions and increased speed variance in human drivers following an
instrumented AV relative to an HDV lead.  From a game-theoretic
standpoint, Millard-Ball~\cite{millardball2018} showed that an AV
which always yields creates a dominant strategy for surrounding road
users to take the available gap, predicting widespread exploitative
behavior at scale.  Our work extends these findings by providing a
naturalistic, paired measurement of the gap-size dimension in cut-in
maneuvers---an aspect prior studies have not directly quantified.

\subsection{Cut-In Detection from Large Datasets}

Wang et al.~\cite{wangXue2019} developed a rule-based cut-in
classifier using lateral displacement thresholds and longitudinal gap
criteria applied to highway NDS data.  Toledo~\cite{toledo2007}
proposed continuous lane-change duration models calibrated on NGSIM
trajectories.  Our detector adapts these frameworks to the AV-centric
coordinate system of WOMD, adding multi-step trajectory buffering and
an agent-type classifier to distinguish AV-targeted from HDV-targeted
events within the same scenario.

% ======================================================================
\section{Dataset}\label{sec:dataset}

\subsection{Waymo Open Motion Dataset}

WOMD~\cite{womd2021} comprises 104{,}000 unique driving scenarios
collected by Waymo's sensor suite across multiple U.S.\ metropolitan
areas.  Each scenario is a 9.1-second segment (91 time steps at
\SI{10}{\hertz}---1 second of historical context followed by 8.1
seconds of future motion) providing 3-D bounding-box tracks for all
road agents (vehicles, pedestrians, cyclists), a high-definition map,
and the AV's kinematic state.  We use the WOMD training partition,
retaining highway and arterial segments where the AV sustains speed
$\geq$\SI{5}{\mps} for at least five consecutive seconds, yielding
\textbf{25{,}906 usable scenarios}.

All positional data are projected into the AV-centric frame
(longitudinal axis aligned with AV heading at the cut-in entry frame),
enabling direct computation of relative kinematics without
map-coordinate artifacts.

% ======================================================================
\section{Methodology}\label{sec:method}

\subsection{Eight-Criterion Cut-In Detector}

A surrounding vehicle triggers a cut-in event when all eight
conditions below are satisfied within a sliding temporal window:

\begin{enumerate}
  \item \textbf{Speed}: surrounding HDV speed $\geq$\SI{2}{\mps}
        throughout the lane-change window.
  \item \textbf{Lateral transition}: signed lateral position crosses
        the lane-center threshold from the adjacent lane into the
        target lane.
  \item \textbf{Longitudinal proximity}: HDV front bumper is within
        \SI{25}{\metre} of the target vehicle at entry.
  \item \textbf{Lane-change duration}: $0.5\leq\Delta t_{lc}\leq6.0$\,s,
        consistent with naturalistic observations~\cite{yang2019}.
  \item \textbf{Relative position}: HDV centroid is ahead of the
        target (positive longitudinal offset) at completion.
  \item \textbf{Lateral velocity}: peak lateral speed
        $\geq$\SI{0.3}{\mps} during the maneuver.
  \item \textbf{No stationarity}: HDV object type is not
        parked/stationary (filtered via the WOMD object-type field).
  \item \textbf{Lane identity}: post-maneuver HDV centroid is within
        \SI{1.5}{\metre} of the target vehicle's lane center.
\end{enumerate}

For \textbf{\HDVtoAV} events, the target is the Waymo AV
(\texttt{sdc\_track\_index}).  For \textbf{\HDVtoHDV} events,
the target is any other tracked vehicle within \SI{75}{\metre} of the
AV in the same scenario, ensuring identical traffic conditions for both
populations.

\subsection{Safety Metrics}

All metrics are computed at the \emph{cut-in entry frame}---the first
frame the HDV's lateral centroid crosses the lane-center threshold:

\begin{itemize}
  \item \textbf{Gap at entry} ($g$): bumper-to-bumper longitudinal
        distance between the cutting-in HDV's rear and the target
        vehicle's front.
  \item \textbf{Time-to-collision} (TTC): $g\,/\,\Delta v$, defined
        only when closing speed $\Delta v{>}0$.
  \item \textbf{Minimum distance} ($d_{\min}$): smallest
        bumper-to-bumper separation over the full lane-change window.
  \item \textbf{Cut-in speed} ($v_{\mathrm{cutin}}$): longitudinal
        speed of the cutting-in HDV at entry.
  \item \textbf{Speed differential} ($\Delta v$):
        $v_{\mathrm{cutin}}-v_{\mathrm{target}}$ at entry.
  \item \textbf{Lane-change duration} ($\Delta t_{lc}$): elapsed time
        from threshold crossing to lane-center stabilization.
  \item \textbf{Lead speed drop}: change in target-vehicle speed over
        the \SI{2}{\second} following entry.
\end{itemize}

\subsection{Statistical Analysis}

All metric distributions are confirmed non-normal by Shapiro-Wilk
tests ($p{<}0.001$).  Hypothesis tests use the \textbf{Mann-Whitney
U test} (two-sided, $\alpha{=}0.05$), with \textbf{Bonferroni
correction} across seven simultaneous metric comparisons
($\alpha_c{=}0.007$).  Effect sizes are reported as \textbf{Cohen's
$d$} (standardized mean difference, pooled SD) and \textbf{Cliff's
$\delta$} (probabilistic effect size).  Bootstrap \SI{95}{\percent}
confidence intervals (10{,}000 resamples) are reported for the primary
outcome (gap at entry); medians for secondary metrics appear in
Table~\ref{tab:descriptive}.

% ======================================================================
\section{Results}\label{sec:results}

\begin{figure*}[!t]
  \centering
  \includegraphics[width=0.98\textwidth]{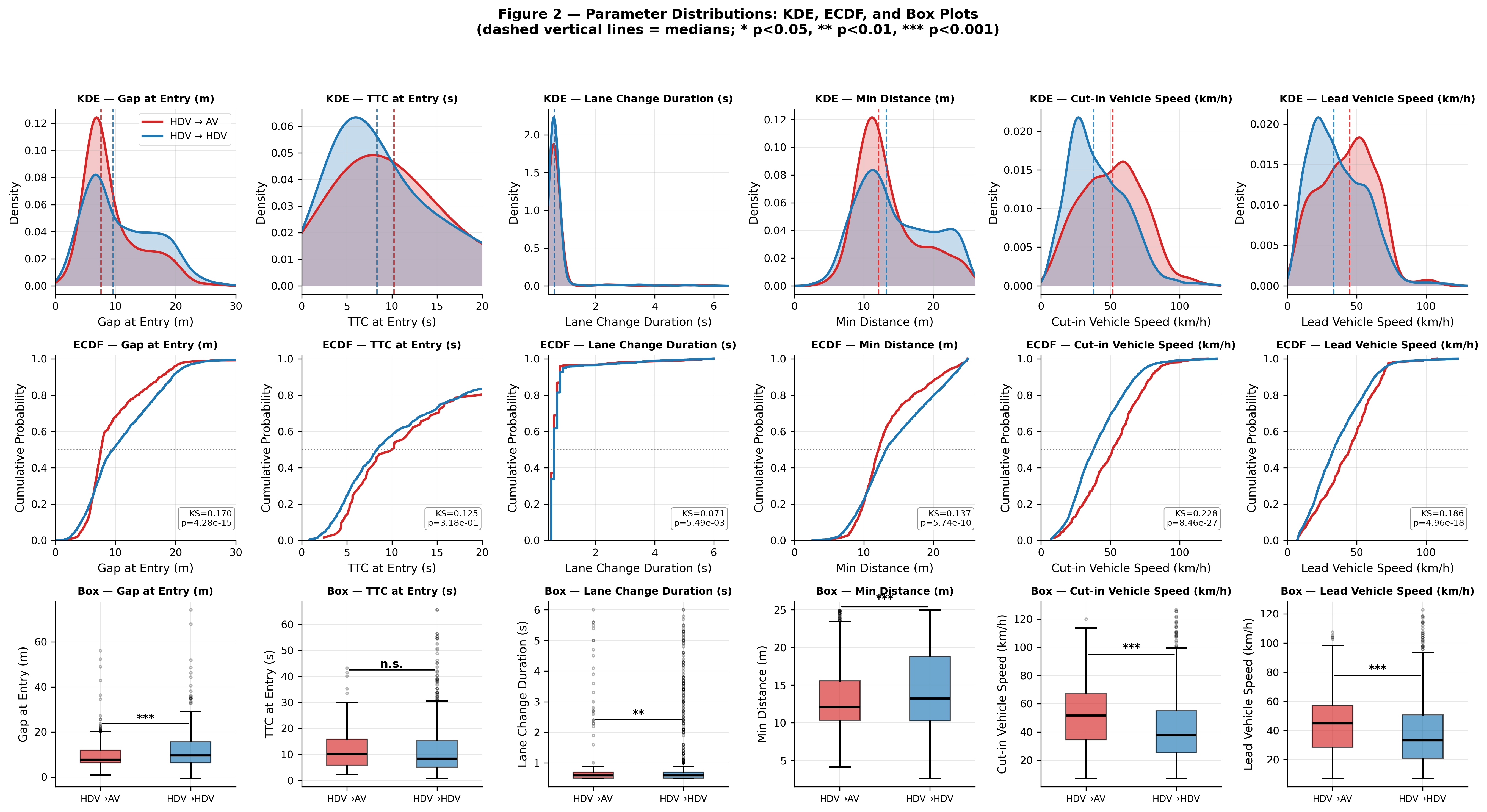}
  \caption{Gap at entry ($d_{LC}$) for \HDVtoAV\ (red) and \HDVtoHDV\
    (blue) cut-in events.
    \textbf{(a)}~Kernel density estimate; dashed vertical lines mark
    group medians; dotted lines indicate the 5\,m and 10\,m risk
    thresholds.
    \textbf{(b)}~Empirical CDF; the AV-targeted curve is consistently
    left of the HDV-targeted curve, corresponding to the 1.99\,m
    median reduction ($p{=}5.76{\times}10^{-8}$, $d{=}{-}0.224$).}
  \label{fig:kde_gap}
\end{figure*}

\subsection{Dataset Summary}

The detector extracted \textbf{706 \HDVtoAV} events (297 from the
left lane, 409 from the right) and \textbf{3{,}172 \HDVtoHDV} events
across the 25{,}906 scenarios.  The 4.5$\times$ higher rate of
HDV-to-HDV events per scenario is consistent with the AV maintaining a
larger following gap, which on average provides fewer cut-in
opportunities at a given traffic density.  Full descriptive statistics
appear in Table~\ref{tab:descriptive}.

% ====== TABLE I ========================================================
\begin{table}[t]
\centering
\caption{Descriptive Statistics: \HDVtoAV\ vs.\ \HDVtoHDV\ Cut-In Events}
\label{tab:descriptive}
\scriptsize
\setlength{\tabcolsep}{2.5pt}
\renewcommand{\arraystretch}{1.15}
\begin{tabularx}{\columnwidth}{@{}L C{0.26\columnwidth} C{0.26\columnwidth}@{}}
\toprule
\textbf{Metric}
  & \textbf{\HDVtoAV} ($n{=}706$)
  & \textbf{\HDVtoHDV} ($n{=}3172$) \\
\midrule
Gap at entry (m)       & 7.58\ [7.43,\,7.80] & 9.57\ [9.14,\,9.99] \\
Min.\ distance (m)     & 12.10               & 13.21 \\
TTC (s)$^{\dagger}$    & 10.22               & 8.34  \\
Cut-in speed (km/h)    & 51.70               & 37.72 \\
Speed diff.\ (km/h)    & 6.90                & 3.83  \\
LC duration (s)        & 0.60                & 0.60  \\
Lead speed drop (km/h) & 2.42                & 1.42  \\
\bottomrule
\end{tabularx}
\vspace{2pt}
{\scriptsize\raggedright
  Gap 95\% CI shown in brackets. $^{\dagger}$TTC requires closing
  speed ${>}0$; 91.4\% of \HDVtoAV\ events have $\Delta v{\leq}0$;
  median computed from 61 valid cases only.\par}
\end{table}

\subsection{Gap at Entry: Primary Outcome}

Figure~\ref{fig:kde_gap} shows the kernel density estimates and
empirical CDFs for gap at entry.  The \HDVtoAV\ distribution is
shifted substantially leftward: the median accepted gap is
\textbf{7.58\,m} (95\%~CI: [7.43,\,7.80]\,m) versus \textbf{9.57\,m}
(95\%~CI: [9.14,\,9.99]\,m) for HDV targets---a \textbf{1.99\,m}
reduction ($U{=}973{,}721$, $p{=}5.76{\times}10^{-8}$,
$d{=}{-}0.224$, Cliff's $\delta{=}{-}0.14$).  This result holds after
Bonferroni correction ($p_c{=}4.6{\times}10^{-7}$).

At the \SI{10}{\metre} risk boundary, \textbf{68.0\%} of \HDVtoAV\
cut-ins fall below this threshold versus \textbf{51.8\%} of \HDVtoHDV\
events ($\chi^2{=}60.5$, $p{<}10^{-13}$).  At the more extreme
\SI{5}{\metre} threshold, the AV-targeted rate is actually lower
(9.1\% vs.\ 14.4\%), suggesting that while drivers routinely accept
moderate-risk gaps in front of the AV, the most extreme gap invasions
are less frequent---plausibly because the AV's reliable deceleration
curtails the conditions under which sub-5\,m gaps would otherwise
persist.

\subsection{Speed Analysis}

Cut-in speed toward the AV (median \SI{51.7}{\kph}) is \textbf{37\%
higher} than toward HDV targets (median \SI{37.7}{\kph};
$p{=}3.14{\times}10^{-31}$, $d{=}0.502$)---a medium-large effect.
The speed differential at entry is 80\% larger as well: \SI{6.90}{\kph}
versus \SI{3.83}{\kph} ($d{=}0.428$, $p{=}3.20{\times}10^{-29}$),
meaning drivers approach the AV at a substantially higher closing speed.

\begin{figure}[t]
  \centering
  \includegraphics[width=\columnwidth]{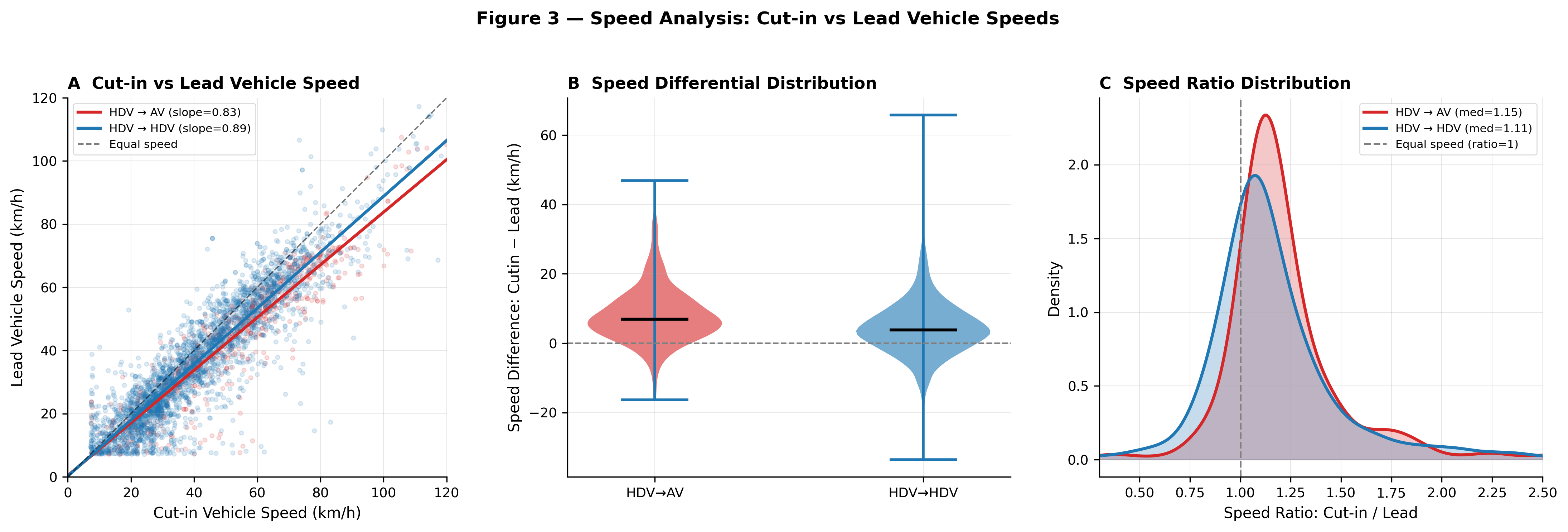}
  \caption{Speed analysis for \HDVtoAV\ (red) and \HDVtoHDV\ (blue)
    events.
    \textbf{(a)}~Cut-in speed vs.\ lead-vehicle speed scatter with
    group regression lines.
    \textbf{(b)}~Violin plot of relative speed differential $\Delta v$
    at entry; AV-targeted events exhibit a 80\% larger differential
    ($d{=}0.428$).
    \textbf{(c)}~KDE of the speed ratio $v_\mathrm{ci}/v_\mathrm{lead}$;
    the AV-targeted distribution peaks above 1, indicating cut-in
    vehicles are faster than their lead target
    ($d{=}0.502$, $p{<}10^{-30}$).}
  \label{fig:speed}
\end{figure}

As Fig.~\ref{fig:speed} illustrates, \textbf{(a)}~AV-targeted events
concentrate at highway speeds (60--100\,km/h), while HDV-targeted events
span a broader urban range; \textbf{(b)}~the speed differential $\Delta v$
is substantially higher for \HDVtoAV\ events; and \textbf{(c)}~the
speed ratio $v_\mathrm{ci}/v_\mathrm{lead}$ confirms that cut-in
vehicles approach the AV at relatively higher speeds.  To verify that
the gap asymmetry is not purely a speed artifact, we repeat the
comparison on speed-matched sub-samples (lead speed
$\in[40,\,65]$\,km/h; $n{=}291$ AV, $n{=}1{,}007$ HDV): the gap
difference persists at \textbf{7.51\,m} vs.\ \textbf{9.25\,m}
($p{<}0.001$), ruling out speed distribution as the sole driver.

\FloatBarrier

\subsection{Full Statistical Summary}

Table~\ref{tab:stats} presents Mann-Whitney test results for all seven
metrics.  Six of seven differ significantly after Bonferroni correction
($\alpha_c{=}0.007$).  TTC ($p{=}0.127$, n.s.) has low statistical
power because valid TTC values exist for only 61 AV-targeted events,
given that 91.4\% of cases have $\Delta v{\leq}0$ at entry.
Lane-change duration ($d{=}{-}0.034$) is statistically significant but
shows no practically meaningful difference, indicating that the lateral
kinematics of the cut-in maneuver are similar regardless of target
type---only the longitudinal gap and speed context differ.

% ====== TABLE II =======================================================
\begin{table}[t]
\centering
\caption{Statistical Comparison: \HDVtoAV\ vs.\ \HDVtoHDV}
\label{tab:stats}
\scriptsize
\setlength{\tabcolsep}{3pt}
\renewcommand{\arraystretch}{1.15}
\begin{tabular}{@{}p{0.40\columnwidth}cccc@{}}
\toprule
\textbf{Metric} & \textbf{Med.\ AV} & \textbf{Med.\ HDV} & \textbf{$p$} & \textbf{$d$} \\
\midrule
Gap at entry (m)       & 7.58  & 9.57  & $5.76{\times}10^{-8}$  & $-$0.224 \\
Min.\ dist.\ (m)       & 12.10 & 13.21 & $1.82{\times}10^{-5}$  & $-$0.201 \\
TTC (s)                & 10.22 & 8.34  & 0.127                  & $+$0.072 \\
Cut-in speed (km/h)    & 51.70 & 37.72 & $3.14{\times}10^{-31}$ & $+$0.502 \\
Speed diff.\ (km/h)    & 6.90  & 3.83  & $3.20{\times}10^{-29}$ & $+$0.428 \\
LC duration (s)        & 0.60  & 0.60  & $1.15{\times}10^{-3}$  & $-$0.034 \\
Lead speed drop (km/h) & 2.42  & 1.42  & $2.72{\times}10^{-4}$  & $-$0.098 \\
\bottomrule
\end{tabular}
\vspace{2pt}
{\scriptsize\raggedright
  Bonferroni-corrected threshold: $\alpha_c{=}0.007$.
  LC duration is statistically significant but practically negligible
  ($|d|{=}0.034$).\par}
\end{table}

\subsection{Joint Risk Space}

\begin{figure}[t]
  \centering
  \includegraphics[width=\columnwidth]{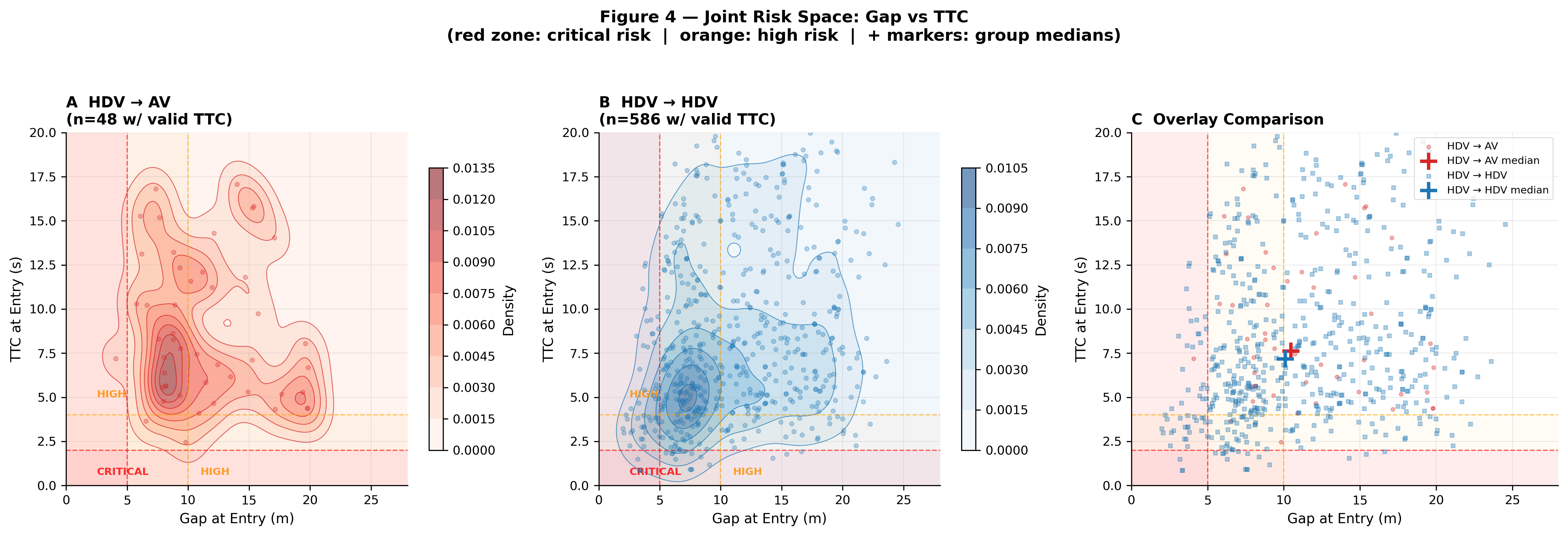}
  \caption{Joint risk space (gap $d_{LC}$ vs.\ TTC) with 2-D KDE
    contours.
    \textbf{(a)}~\HDVtoAV\ density map; \textbf{(b)}~\HDVtoHDV\
    density map; \textbf{(c)}~overlay scatter with group medians
    (diamonds).  Shaded regions mark critical ($d{<}5$\,m,
    TTC\,$<$\,1.5\,s) and high-risk ($d{<}10$\,m, TTC\,$<$\,3\,s)
    zones.  \HDVtoAV\ events cluster toward the lower-left;
    \HDVtoHDV\ events occupy a wider, lower-risk region.}
  \label{fig:risk_space}
\end{figure}

Figure~\ref{fig:risk_space} overlays 2-D KDE contours in the
gap--TTC plane.  \textbf{Panel~(a)} shows \HDVtoAV\ mass concentrating
in the lower-left quadrant (small gap, moderate TTC). \textbf{Panel~(b)}
shows \HDVtoHDV\ events dispersed toward larger gaps and higher TTC.
\textbf{Panel~(c)} combines both groups with group medians (diamonds);
the \HDVtoAV\ median falls in the moderate-risk zone, while the
\HDVtoHDV\ median lies in the low-risk region.  The larger post-cut-in
speed drop for AV-targeted events (\SI{2.42}{\kph} vs.\ \SI{1.42}{\kph},
$p{<}0.001$) confirms that the AV decelerates actively to restore a
safety margin---a compensatory response that is less consistent in
HDV-to-HDV interactions.

\FloatBarrier

\subsection{Severity Classification}

Because TTC is undefined for 91.4\% of \HDVtoAV\ events (the AV's
speed equals or exceeds the cutter's at entry), we classify severity
using gap at entry alone---a metric defined for every event:

\begin{itemize}
  \item \textbf{Critical}: gap\,$<$\,\SI{5}{\metre}
  \item \textbf{Moderate}: \SI{5}{\metre}\,$\leq$\,gap\,$<$\,\SI{10}{\metre}
  \item \textbf{Low}: gap\,$\geq$\,\SI{10}{\metre}
\end{itemize}

\begin{figure}[t]
  \centering
  \includegraphics[width=\columnwidth]{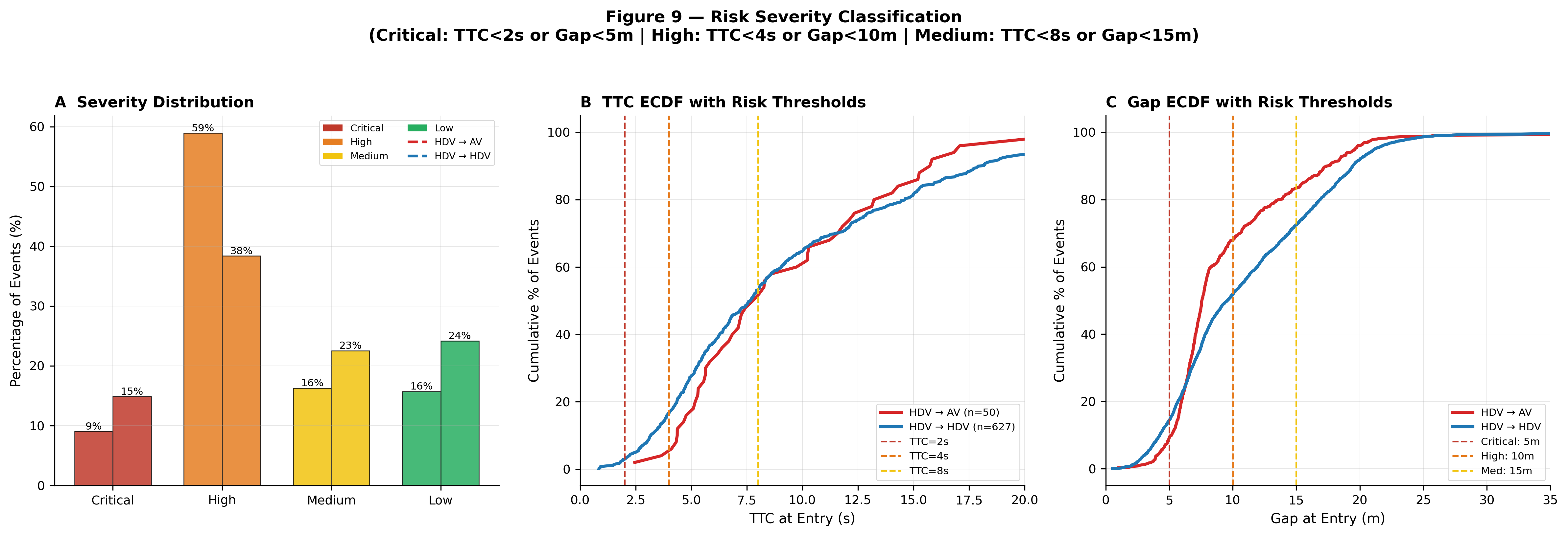}
  \caption{Severity classification by gap-at-entry threshold.
    \textbf{(a)}~Proportion of events in each severity class for both
    groups; AV-targeted events show a higher Moderate share (58.9\%
    vs.\ 37.4\%), while HDV-targeted events have more Critical cases
    (14.4\% vs.\ 9.1\%).
    \textbf{(b)}~Empirical CDF of TTC (events with valid TTC only);
    dotted/dash-dot lines mark the 1.5\,s and 3.0\,s thresholds.
    \textbf{(c)}~Empirical CDF of gap $d_{LC}$ with the 5\,m and
    10\,m risk boundaries; the \HDVtoAV\ curve lies consistently to
    the left, confirming higher aggregate risk exposure.}
  \label{fig:severity}
\end{figure}

Critical events account for \SI{9.1}{\percent} of AV-targeted cut-ins
versus \SI{14.4}{\percent} of HDV-targeted events.  Moderate events
are substantially more prevalent among AV targets:
\SI{58.9}{\percent} versus \SI{37.4}{\percent}.  Combined, 68.0\% of
AV-targeted cut-ins reach critical or moderate severity, compared with
51.8\% of HDV-targeted events---a 16.2 percentage-point higher
exposure to elevated-risk conditions (Fig.~\ref{fig:severity}).

\FloatBarrier

\subsection{Case Replay: Example Cut-In Scenario}
\label{sec:replay}

\begin{figure}[t]
  \centering
  \includegraphics[width=\columnwidth]{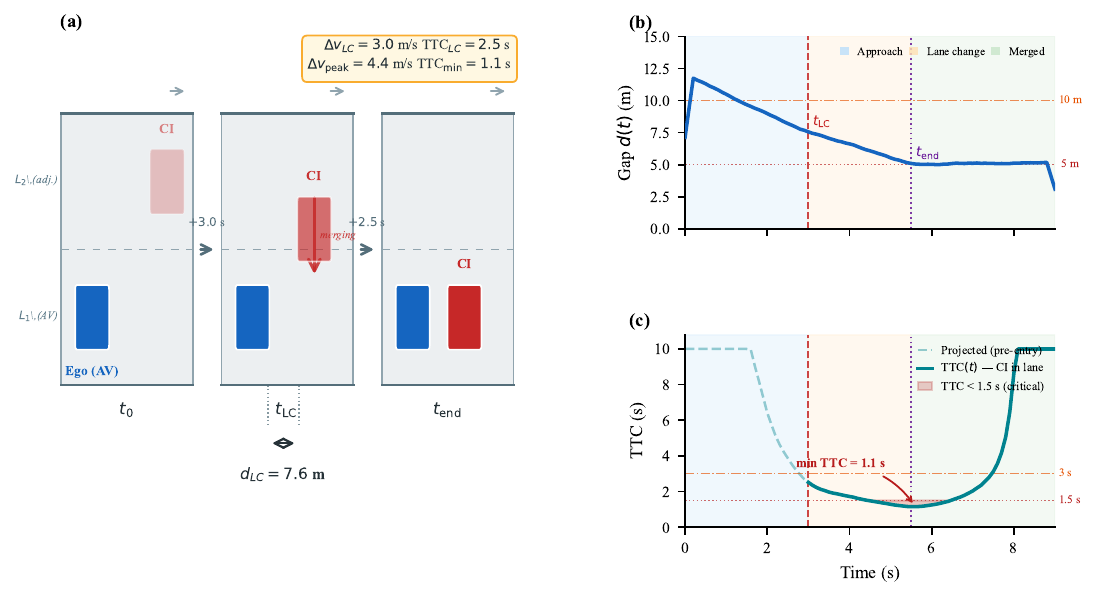}
  \caption{Case replay of a representative \HDVtoAV\ cut-in event.
    \textbf{(a)}~Bird's-eye plan view: ego AV (blue, fixed) and
    three ghost positions of the cut-in HDV (red) at $t_0$,
    $t_\mathrm{LC}$, and $t_\mathrm{end}$; dashed curve shows the
    lateral trajectory; $d_\mathrm{LC}$ is the bumper-to-bumper gap
    at lane-change entry.  The info box reports peak approach speed
    $\Delta v_\mathrm{peak}$ and minimum TTC.
    \textbf{(b)}~Gap $d(t)$ time-series with phase shading
    (free-flow / lane-change / merged) and risk thresholds
    (10\,m dash-dot; 5\,m dotted).
    \textbf{(c)}~TTC$(t)$; dashed line shows projected TTC during the
    pre-entry phase (CI in adjacent lane); minimum TTC of 1.1\,s
    is reached near $t_\mathrm{end}$ and recovers post-merge.}
  \label{fig:replay}
\end{figure}

To ground the aggregate statistics in a concrete event,
Fig.~\ref{fig:replay} replays a representative \HDVtoAV\ cut-in
scenario from WOMD.  The cutting-in HDV begins in the adjacent lane
($L_2$) and completes a $\Delta t_{lc}{=}2.5$\,s lane change into
$L_1$ ahead of the AV.  Table~\ref{tab:replay} summarises the
kinematic state at each key instant.

\begin{table}[h]
\centering
\caption{Kinematic State at Key Instants -- Representative \HDVtoAV\ Scenario}
\label{tab:replay}
\scriptsize
\setlength{\tabcolsep}{3.5pt}
\renewcommand{\arraystretch}{1.15}
\begin{tabular}{@{}lccc@{}}
\toprule
\textbf{Instant} & \textbf{Gap (m)} & \textbf{$\Delta v_\mathrm{app}$ (m/s)} & \textbf{TTC (s)} \\
\midrule
$t_0$ \ (LC onset)          & 12.0 & $<$0 (CI faster) & --- \\
$t_\mathrm{LC}$ (LC entry)  & 7.6  & $+$3.0           & 2.5 \\
$t_\mathrm{end}$ (merged)   & 5.0  & $+$4.4           & 1.1 \\
$t_\mathrm{end}{+}3\,\mathrm{s}$ (recovery) & 5.5 & $\approx$0 & ${>}3$ \\
\bottomrule
\end{tabular}
\vspace{2pt}
{\scriptsize\raggedright
  $\Delta v_\mathrm{app}{=}v_\mathrm{ego}{-}v_\mathrm{CI}$;
  positive values indicate the ego vehicle is closing on the cut-in
  vehicle.  TTC reported only when $\Delta v_\mathrm{app}{>}0$.\par}
\end{table}

Panel~(a) of Fig.~\ref{fig:replay} shows the bird's-eye plan view.
At $t_0$, the CI vehicle is fully in $L_2$ with a 12\,m standoff gap.
At $t_\mathrm{LC}$, it has crossed the lane boundary with $d_{LC}{=}7.6$\,m---
closely matching the population median (7.58\,m)---and is beginning
to brake.  By $t_\mathrm{end}$, the CI vehicle is fully merged with a
5.0\,m gap, classifying this event as \emph{Moderate} severity.  The
peak approach speed rises to 4.4\,m/s as the CI vehicle decelerates
sharply within the AV's lane.

Panels~(b) and (c) confirm the temporal dynamics.  The gap $d(t)$
drops monotonically from 12\,m to 5.0\,m over the 2.5\,s merge
window before stabilising.  TTC, undefined pre-entry (CI is faster),
drops sharply after $t_\mathrm{LC}$ as the CI vehicle brakes, reaching
a minimum of 1.1\,s at $t_\mathrm{end}$.  Within 3\,s of
merge completion the AV's deceleration restores TTC above 3\,s---
consistent with the population-level observation that AV-targeted
events exhibit a \SI{2.42}{\kph} post-entry speed drop versus
\SI{1.42}{\kph} for HDV-targeted events (Table~\ref{tab:descriptive}).

\FloatBarrier

% ======================================================================
\section{Discussion}\label{sec:discuss}

\subsection{Exploitation vs.\ Implicit Recalibration}

Two non-exclusive mechanisms can account for the observed asymmetry.
The first is \emph{purposeful exploitation}: knowing that an AV will
not contest a cut-in through acceleration or social signaling, some
drivers deliberately accept gaps they would not take in front of an
attentive human driver.  This behavior is predicted by game-theoretic
analysis~\cite{millardball2018} and confirmed in field
experiments~\cite{soni2022}.  The 80\% larger speed differential
($d{=}0.428$) supports this interpretation: drivers approach the AV at
higher closing speeds, implying an expectation of accommodation rather
than resistance.

The second mechanism is \emph{implicit recalibration}: drivers
subconsciously reduce their gap threshold because the AV's smooth,
reliable deceleration lowers perceived risk.  Under this account the
driver is not gaming the AV but rather updating their internal risk
model accurately, given the AV's consistent braking
behavior~\cite{wen2022,rahmati2019}.

From an engineering perspective both mechanisms converge on the same
outcome---more frequent cut-ins at shorter gaps---requiring the AV to
plan for an interaction distribution that diverges from HDV-calibrated
baselines.

\subsection{Implications for AV Motion Planning}

Current AV safety envelopes are typically calibrated on HDV-to-HDV
gap-acceptance distributions~\cite{yang2019}, where the median
accepted gap is approximately 9.57\,m.  Our results show that 68.0\%
of real-world AV-targeted cut-ins fall below the \SI{10}{\metre}
boundary---a gap range where HDV-trained threat-assessment models may
assign low priority, yet one that represents the majority of the AV's
actual exposure.  An AV whose precautionary deceleration is triggered
at the HDV-calibrated median (9.57\,m) may not activate at
7.58\,m despite the genuine risk, because that gap falls inside its
trained ``normal'' range.

We recommend that AV highway motion planners apply an additional
longitudinal buffer of at least \SI{2}{\metre} in lane-adjacent
scenarios---equivalent to the empirical gap shift measured here---to
better align the AV's internal model with the gap distribution it
actually encounters in mixed traffic.

\subsection{Implications for Traffic Simulation}

Microscopic traffic simulators (SUMO, VISSIM, AIMSUN) draw
gap-acceptance parameters from HDV-only calibration studies.  In
mixed-traffic simulations, this leads to underestimation of both the
frequency and severity of AV-targeted cut-in events.  The empirical
distributions reported here can serve as direct calibration targets for
AV-specific gap-acceptance submodels.

\subsection{Limitations}

This study has several limitations.  The WOMD scenarios are collected
in specific U.S.\ cities (San Francisco, Phoenix) and may not
generalize to other driver populations or road environments.  Driver
intent cannot be observed directly; the target classification is based
on trajectory geometry, not on gaze or stated decision processes.  The
HDV-to-HDV population includes vehicles up to \SI{75}{\metre} from the
AV, which may introduce traffic-density confounds not fully eliminated
by speed matching.  The AV's own following policy---maintaining a
systematically larger gap---may also create the physical space that
enables shorter accepted gaps in front of it, an endogenous feedback
difficult to disentangle from genuine driver intent using observational
data alone.  Future work using propensity-score matching on
leader-type covariates, or instrumented HDV comparators, would allow
stronger causal attribution.

% ======================================================================
\section{Conclusion}\label{sec:conclude}

This paper presents the first large-scale, naturalistic comparison of
cut-in gap acceptance when the lead vehicle is an autonomous versus a
human-driven vehicle.  Drawing on 706 \HDVtoAV\ and 3{,}172 \HDVtoHDV\
events from the Waymo Open Motion Dataset, we find:

\begin{enumerate}
  \item Human drivers accept a \textbf{1.99\,m smaller median gap}
        when cutting in front of the AV (7.58 vs.\ 9.57\,m,
        $p{<}10^{-7}$, $d{=}{-}0.224$), an effect that persists
        under speed-matched resampling.
  \item Cut-in speeds toward the AV are \textbf{37\% higher}
        (51.7 vs.\ 37.7\,km/h, $d{=}0.502$), with an 80\% larger
        closing speed differential ($d{=}0.428$).
  \item \textbf{68.0\%} of AV-targeted cut-ins fall below the 10\,m
        gap boundary, versus 51.8\% for HDV targets
        ($\chi^2{=}60.5$, $p{<}10^{-13}$).
  \item The AV's active deceleration response (2.42 vs.\ 1.42\,km/h
        speed drop post-cut-in) provides a compensating safety factor
        absent in HDV-to-HDV interactions, as confirmed by the case
        replay (Section~\ref{sec:replay}).
\end{enumerate}

The asymmetry is statistically robust and practically significant.
It calls for AV-specific calibration of gap-acceptance models in
motion planning and simulation, and raises a regulatory question about
whether AVs should exhibit limited behavioral variability to reduce
systematic exploitation without compromising collision avoidance.

% ======================================================================
% Acknowledgment (OMIT for blind review)
\ifblindreview
  % intentionally omitted
\else
  \section*{Acknowledgment}
  The authors thank Waymo for making the Open Motion Dataset publicly
  available.  [Funding acknowledgment to be added before camera-ready
  submission.]
\fi

% ======================================================================
\balance
\bibliographystyle{IEEEtran}

\end{document}